\documentclass[journal]{IEEEtran}
\IEEEoverridecommandlockouts
\def\IEEEtitletopspace{18pt}
\usepackage{cite}
\usepackage{amsmath,amssymb,amsfonts}
\usepackage{enumitem}
\usepackage{graphicx}
\usepackage{array}    
\usepackage{booktabs} 
\usepackage[font=small,labelfont=bf]{caption}
\usepackage{makecell} 
\usepackage{comment}
\usepackage{booktabs}
\usepackage{siunitx}
\usepackage{pifont}            
\usepackage{empheq}

\usepackage{adjustbox} 
\usepackage{url}
\usepackage{float}
\usepackage{threeparttable} 
\usepackage{graphicx}
\usepackage{subcaption}
\usepackage{multirow} 
\usepackage{flushend}
\usepackage[capitalise]{cleveref}
\usepackage{textcomp}
\usepackage{amsmath}      
\usepackage{algorithm}    
\usepackage{algpseudocode} 
\usepackage{float}        
\usepackage{xcolor,colortbl}
\usepackage[top=54pt, left=54pt, right=54pt, bottom=54pt]{geometry}%
\def\BibTeX{{\rm B\kern-.05em{\sc i\kern-.025em b}\kern-.08em
    T\kern-.1667em\lower.2ex\hbox{E}\kern-.125emX}}

\definecolor{Gray}{gray}{0.75}
\definecolor{Grey}{gray}{0.92}

\newcolumntype{a}{>{\columncolor{Grey}}c}

\title{Learning Agile Tensile Perching for Aerial Robots from Demonstrations}

\begin{document}
\bstctlcite{IEEEexample:BSTcontrol}
\date{}

\author{
Kangle Yuan$^{1,*}$, Atar Babgei$^{1,4*}$, Luca Romanello$^{1,2}$, Hai-Nguyen Nguyen$^{6}$, \\ Ronald Clark$^{3}$, Mirko Kovac$^{1,4}$, Sophie F. Armanini$^{1,2}$, Basaran Bahadir Kocer$^{1,5}$  \\

\thanks{
\par $^{1}$Aerial Robotics Laboratory, Imperial College London.
\par $^{2}$eAviation Laboratory, TUM School of Engineering and Design, Munich.
\par $^{3}$Department of Computer Science, University of Oxford.
\par $^{4}$Laboratory of Sustainability Robotics, EMPA, Dübendorf, Switzerland.
\par $^{5}$School of Civil, Aerospace and Design Engineering, University of Bristol.
\par $^{6}$ School of Science, Engineering and Technology, RMIT University Vietnam.
\par $^{*}$The first and second authors contributed equally to this work.
}
}


\maketitle
\begin{abstract}

Perching on structures such as trees, beams, and ledges is essential for extending the endurance of aerial robots by enabling energy conservation in standby or observation modes. A tethered tensile perching mechanism offers a simple, adaptable solution that can be retrofitted to existing robots and accommodates a variety of structure sizes and shapes. However, tethered tensile perching introduces significant modelling challenges which require precise management of aerial robot dynamics, including the cases of tether slack \& tension, and momentum transfer. Achieving smooth wrapping and secure anchoring by targeting a specific tether segment adds further complexity. In this work, we present a novel trajectory framework for tethered tensile perching, utilizing reinforcement learning (RL) through the Soft Actor-Critic from Demonstrations (SACfD) algorithm. By incorporating both optimal and suboptimal demonstrations, our approach enhances training efficiency and responsiveness, achieving precise control over position and velocity. This framework enables the aerial robot to accurately target specific tether segments, facilitating reliable wrapping and secure anchoring. We validate our framework through extensive simulation and real-world experiments, and demonstrate effectiveness in achieving agile and reliable trajectory generation for tensile perching.

\end{abstract}

\section{Introduction}
\label{section:introduction}

Aerial robots have shown promise in applications like infrastructure inspection\cite{shakhatreh_unmanned_2019}, and forestry observation \cite{nguyen2024crash,pritchard2025forestvo,bates2025leaf,kocer_forest_2021,kirchgeorg2024eprobe,geckeler2025field} by reaching hard-to-access areas with minimal human interventions. However, these applications are often limited by battery constraints, operational noise, and the need for stable positioning to collect accurate data. Perching on structures provides an effective solution, allowing aerial robots to conserve energy, reduce noise by stopping propellers, and maintain stability for precise data capture.

A tethered tensile perching \cite{zhang_spidermav_2017,braithwaite_tensile_2018,nguyen_passively_2019,hauf_learning_2023,lan_aerial_2024}, offers a simple and adaptable alternative, compared to the other perching technologies such as mechanical grippers and hooks \cite{zheng_albero_2024, roderick_bird-inspired_2021,li2025treecreeper}, ceiling effect \cite{zou_perch_2023}, and special landing materials either adhesive \cite{liu_electrically_2023, daler_perching_2013} or magnetic \cite{ji_real-time_2022, garcia-rubiales_magnetic_2019}. By using a tether to wrap around structures, tethered tensile perching can be retrofitted to existing aerial robots and is less dependent on perching target (e.g. branch-like structure) size or shape, making it suitable for diverse environments.

\begin{figure}[t!]
    \centering
    \includegraphics[width=1.0\linewidth]{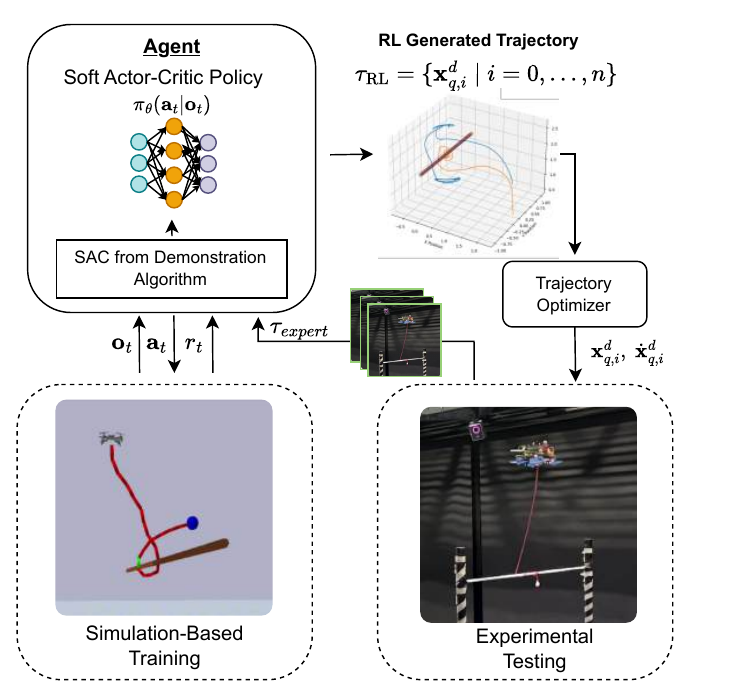}
    \caption{Overview of the learning-based agile tensile perching strategy for aerial robots, combining dynamic system modelling with trajectory learning from demonstrations.}
    \label{fig:overview}
\end{figure}

However, tethered tensile perching introduces complex challenges, as the addition of a tether and perching weight requires modelling the aerial robot with a suspended perching weight, creating a multifaceted control problem involving free flight dynamics, tether slack and tension, momentum transfer, and other nonlinear interactions \cite{palunko_trajectory_2012,guerrero_attenuation_2017}. Additionally, successful perching demands precise control over both position and velocity to target a specific segment of the tether, allowing the perching weight to wrap smoothly around the structure and generate sufficient momentum for secure anchoring \cite{hauf_learning_2023}.

\begin{figure*}[!t]
    \centering
    \begin{subfigure}{0.35\textwidth}
        \centering
        \includegraphics[width=\linewidth]{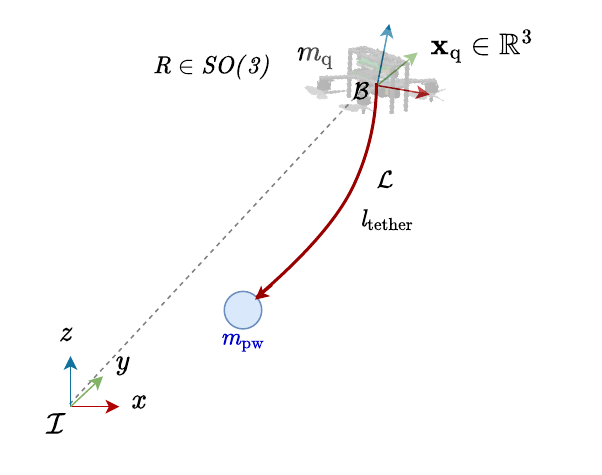}
        \caption{}
        \label{fig:coordinate}
    \end{subfigure}%
    \hfill
    \begin{subfigure}{0.65\textwidth}
        \centering
        \includegraphics[width=\linewidth]{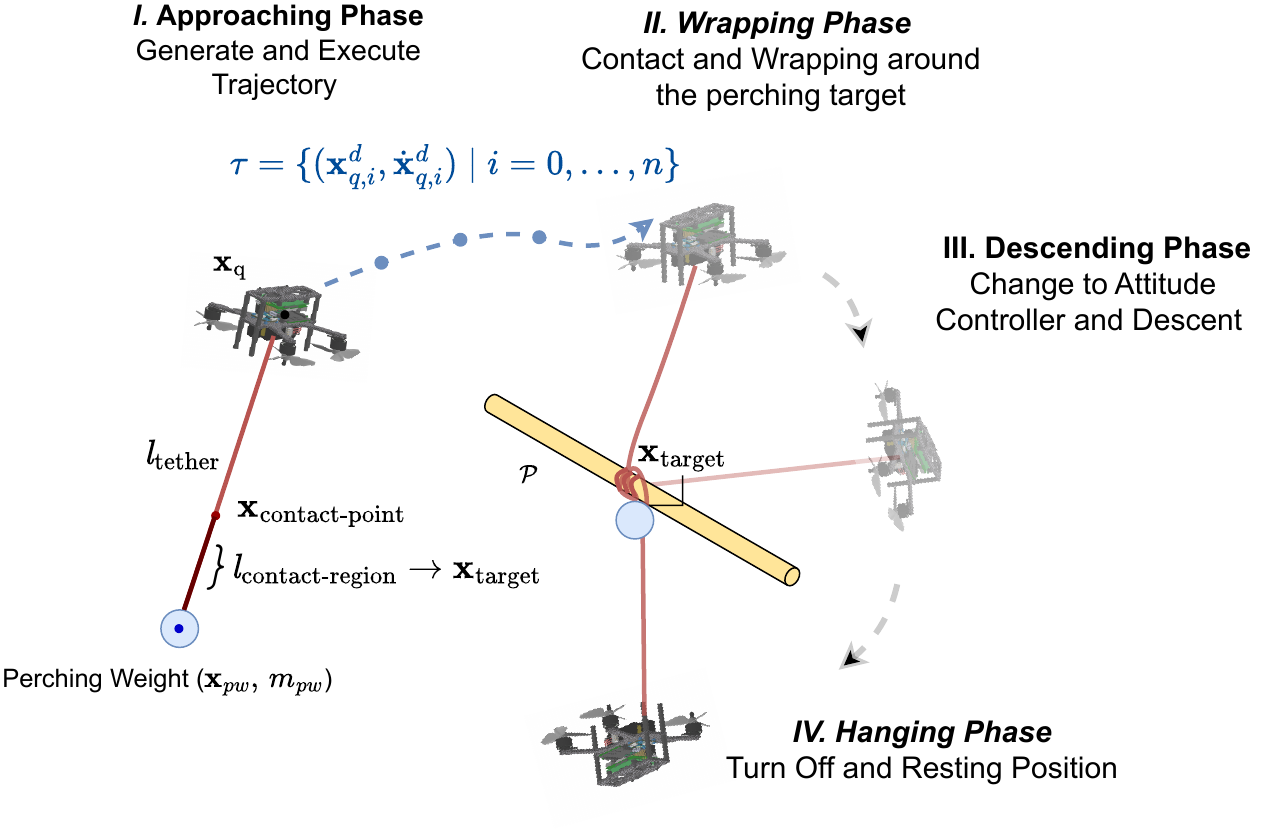}
        \caption{}
        \label{fig:whole-sequences}
    \end{subfigure}
    \caption{(a) 
The quadrotor with a perching mechanism (tether cable and perching weight) . (b) The four-phase sequence of the perching manoeuvre: (I) Approaching Phase, in which the robot generates and executes its trajectory toward the perching target; (II) Wrapping Phase, where the tether makes contact and wraps securely around the perching target; (III) Descending Phase, involving the transition to attitude control in real-world experiments only for a stable descent; and (IV) Hanging Phase, where the robot powers off and stabilizes in the final resting position in real-world experiments.}
    \label{fig:combined}
\end{figure*}

Model-based control approaches face challenges with these requirements. For example, optimisation-based controllers such as MPC are effective in predictable conditions \cite{kocer2019aerial}. Still, they lack adaptability due to the changes in tether dynamics in slack and taut cases and variable tether-quadrotor interactions. On the other hand, model-free approaches like reinforcement learning (RL) offer promising solutions for complex trajectory generation and control tasks. RL has proven effective for challenging tasks like aggressive perching on walls \cite{lee2021low} and adaptive planning under uncertainty \cite{ruckin2022adaptive}, demonstrating its potential for dynamic, nonlinear control tasks. However, existing RL-based approaches have not yet addressed the specific challenges of tethered tensile perching.

RL training is resource-intensive, especially when managing complex state and action spaces \cite{rajeswaran2017learning}. Applying RL to tethered tensile perching adds challenges in exploration and learning efficiency. The precise targeting and complex tether dynamics create a sparse reward structure, where successful perching events are rare and offer limited feedback. Random exploration is ineffective, as the agent is unlikely to discover successful behaviours without guidance. The control inputs and the need for long-horizon planning to manage the tether amplify the difficulty. Additionally, physical interactions, especially during the transition after initial contact, make the environment sensitive to small errors. This complexity makes learning through trial and error challenging. To address these issues, incorporating demonstrations into the RL framework can guide exploration \cite{nair2018overcoming}. By leveraging expert trajectories that showcase successful perching and tether management, the agent can focus on promising areas of the state space by accelerating the learning process and improving overall performance in this demanding task. 

In this study, we present a novel framework for generating agile trajectories for tensile perching using RL enhanced with demonstration data $\tau_{expert}$ collected from a human expert to expedite training, as illustrated in Figure \ref{fig:overview}. While our former work only focused on the last stage of upside-down trajectories to shut down the motors \cite{hauf_learning_2023}, we are releasing a full perching framework with the following main highlights:

\begin{itemize}
    \item We develop a framework for precise tensile perching trajectories by separating position and velocity control. The RL model generates the position trajectory, while an optimization step computes the velocity profile, ensuring correct dynamics for secure wrapping around target structures.
    
    \item We implement the Soft Actor-Critic from Demonstration (SACfD) algorithm, using demonstration data, including suboptimal samples, to improve training efficiency and performance for tensile perching.
    
    \item We validate our framework through real-world experiments with a custom aerial robot, demonstrating improved perching precision, robustness, and adaptability to diverse environments.

    \item We provide an open-source simulation environment in PyBullet\footnote{https://github.com/AerialRoboticsGroup/agile-tethered-perching} for the tethered aerial robot, capturing essential nonlinear dynamics and serving as a testbed for trajectory generation and control strategies.
\end{itemize}


\section{Methodology} 
\label{section:methodology}


Figure~\ref{fig:coordinate} shows the setup: a quadrotor of mass~$m_{\mathrm q}$ carries a perching weight~$m_{\mathrm{pw}}$ on a tether of fixed length~$l_{\text{tether}}$.  
Traditional suspended-payload control tries to stop the load from swinging.   Tensile perching, by contrast, must swing it on purpose.  
Here the policy’s job is to find a flight trajectory that drives the weight to the target~$\mathcal{P}$ and lets it wrap around on impact, completing the perch.


We introduce a five-stage sequence that integrates trajectory planning, contact handling, and attitude-controlled descent (only in real-world experiments) to achieve stable perching manoeuvres, as shown in Figure \ref{fig:whole-sequences}. Using the SACfD algorithm, our model learns optimal policies that balance computational efficiency with physical accuracy. SACfD extends the Soft Actor-Critic (SAC) algorithm by incorporating demonstration data into the learning process, which accelerates early-stage learning and enhances policy performance \cite{nematollahi2022robot}. These demonstrations \( \tau_{expert}\) (see Figure \ref{fig:overview}) were collected from human operator within a laboratory environment.

\subsection{Frame Definition}
\label{subsection:dynamics-modelling}

We employ two coordinate frames: the inertial frame $\mathcal{I}$ (aligned with gravity) and the quadrotor body frame $\mathcal{B}$, which translates and rotates with the vehicle (see Figure~\ref{fig:coordinate}). The quadrotor’s attitude relative to $\mathcal{I}$ is $R\in SO(3)$, its center‐of‐mass position is $\mathbf{x}_q\in\mathbb{R}^3$, and its mass is $m_q$. We model a perching payload as a point mass $m_{pw}$ suspended by a massless tether; its position is $\mathbf{x}_{pw}\in\mathbb{R}^3$, and the length of tether $\mathcal{L}$ is $\ell_{\mathrm{tether}}$. The desired perching target $\mathcal{P}$ is located at $\mathbf{x}_{\mathrm{target}}\in\mathbb{R}^3$.

\subsection{Reinforcement Learning and Policy Optimization for Perching Tasks}
\label{subsection:rl-overview}

\begin{figure}[t]
    \centering
    \includegraphics[width=1\linewidth]{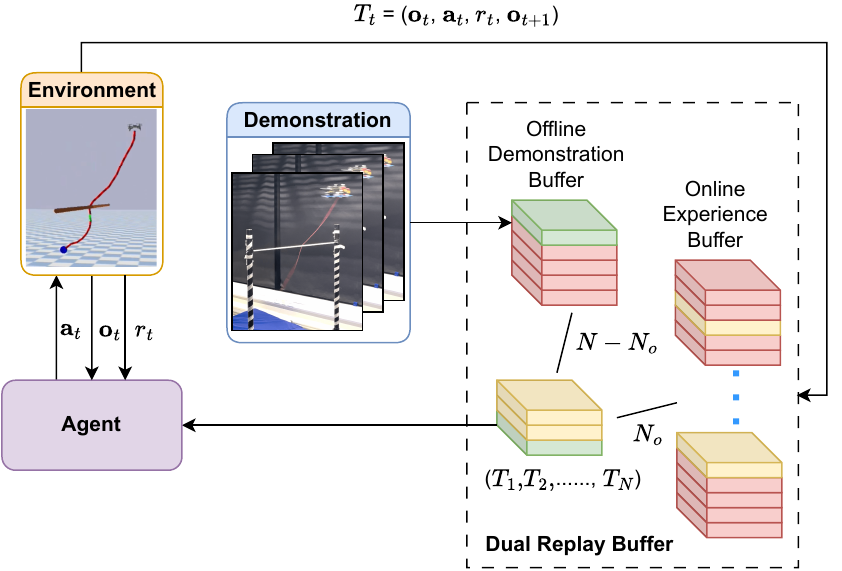}
    \caption{SACfD algorithm training system which combines reinforcement learning (SAC) with learning from demonstrations. The agent interacts with the environment by receiving observation (\(\mathbf{o}_t\)), taking actions (\(\mathbf{a}_t\)), and obtaining rewards (\(r_t\)). Transitions \((T_t = (\mathbf{o}_t, \mathbf{a}_t, r_t, \mathbf{o}_{t+1})\)) are stored in a dual replay buffer, which consists of an offline demonstration buffer and an online experience buffer. During training, a total batch size of \(N\) is sampled from both buffers by leveraging demonstration data and online experiences to enhance policy learning and performance.}
    \label{fig:sacfd-algo}
\end{figure}

We adopt RL to enable the quadrotor to learn robust perching policies by maximizing cumulative rewards through iterative environment interactions. At each time step \( t \), the cumulative reward \( R(\mathbf{o}_t, \mathbf{a}_t) \) is calculated as:
\begin{equation}
R(\mathbf{o}_t, \mathbf{a}_t) = \sum_{k=0}^{T} \gamma^k r_{t+k+1},
\end{equation}
where \( r_{t+k+1} \) represents the reward received \( k+1 \) time steps into the future, and \( \gamma \) is a discount factor balancing the importance of immediate versus future rewards.

At each time step \(t\), the agent receives an observation
\[
  \mathbf{o}_t
  = \bigl\{\,\mathbf{x}_{q}(t),\;w(t),\;\eta(t)\bigr\},
\]
where
\begin{itemize}
  \item \(\mathbf{x}_{q}(t)\in\mathbb{R}^3\) is the quadrotor’s current position in the inertial frame \(\mathcal{I}\),
  \item \(w(t)\in\mathbb{N}\) is the current wrap count of the tether around the perching target and
  \item \(\eta(t)\in[0,1]\) is the fraction of the perching task completed (0 = start, 1 = fully wrapped and hanged, or reached the step limit).
\end{itemize}

The action issued by the policy at time \(t\) is the next desired waypoint,
\[
  \mathbf{a}_t = \mathbf{x}_{q,i}^{d} \;\in\;\mathbb{R}^3,
\]
where \(i\) indexes the waypoint in the planned sequence for that episode.  In practice, the policy outputs a 3-D position offset or absolute waypoint in \(\mathcal{I}\), which the onboard controller then tracks. 

We chose the SAC algorithm from the Stable Baselines3 (SB3) framework \cite{raffin_stable-baselines3_2021} due to its effective balance between exploration and exploitation via entropy regularization. SAC optimizes two Q-functions, \( Q_{\phi_1}(\mathbf{o}_t, \mathbf{a}_t) \) and \( Q_{\phi_2}(\mathbf{o}_t, \mathbf{a}_t) \), minimizing overestimation bias by selecting the lower of the two:
\begin{equation}
\small
J(\theta) = \mathbb{E}_{\mathbf{o}_t \sim \rho^{\pi}, \mathbf{a}_t \sim \pi_{\theta}} \left[ \alpha \log \pi_{\theta}(\mathbf{a}_t | \mathbf{o}_t) - \min_{i=1,2} Q_{\phi_i}(\mathbf{o}_t, \mathbf{a}_t) \right],
\end{equation}
where \( \alpha \) is the entropy coefficient promoting exploratory behaviour.

To enhance learning, we extend SAC into SACfD, which incorporates offline demonstrations from human piloting to accelerate training as illustrated in Figure \ref{fig:sacfd-algo}. SACfD uses a hybrid replay buffer consisting of both online experiences from live interaction with the environment and offline expert demonstrations. At each training iteration, a batch of \(N_o\) samples is drawn from the online buffer and \(N - N_o\) from the offline buffer:
\begin{equation}\label{eq:sampling}
    N_o = \min \left( \left\lfloor \frac{N}{\lambda} \right\rfloor, 10^6 \right)
\end{equation}
where \( N \) represents the total batch size and \(\lambda\) is the weighting constant.

This dual-buffer approach facilitates the learning from both resources. Online experiences, updated through live interaction, keep the agent adaptive to the current environment, while offline experiences, derived from collected expert trajectory data, accelerate early learning for initial guidance. Pre-training on offline data further expedites learning, enabling the agent to generalize effectively across diverse perching scenarios. During training, deterministic validation ensures robust performance by saving the best-performing policies.

\subsection{Reward Function Design}
\label{subsection:reward-function}

\begin{figure}[!t]
    \centering
    \includegraphics[width=1.0\linewidth]{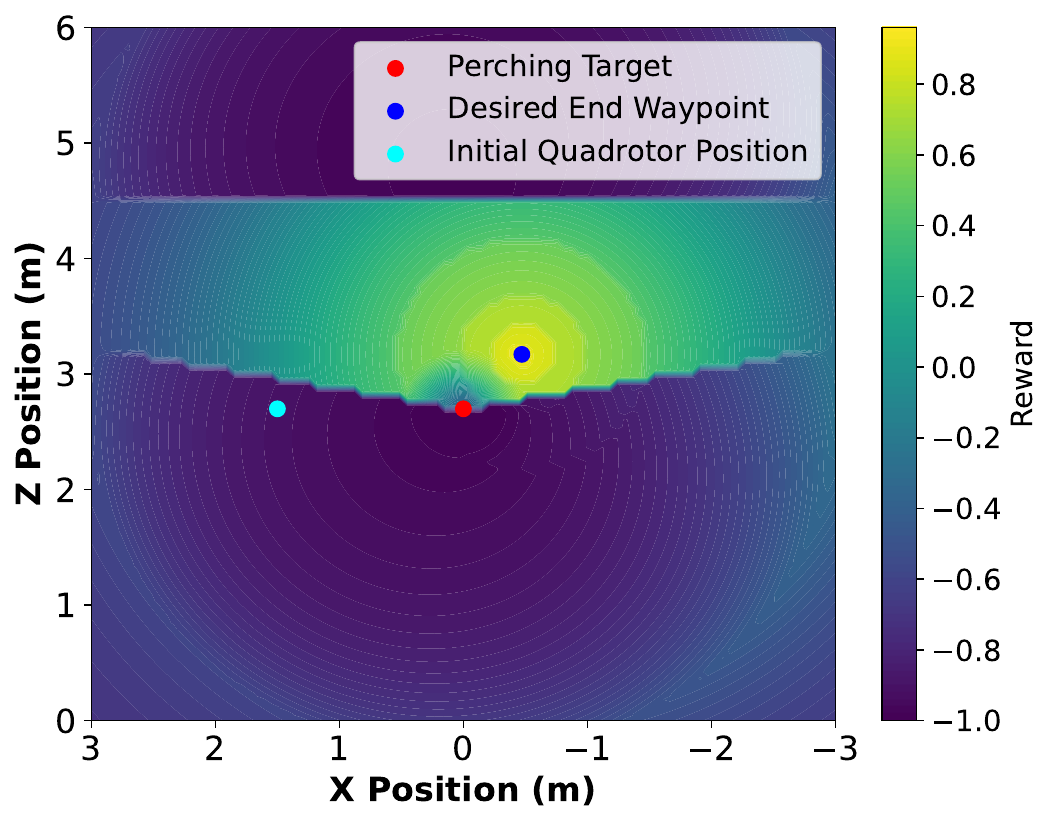}
    \caption{Visualization of the reward function \(\mathbf{R_{\text{approach}}}\) 
  in the X-Z plane. The heatmap illustrates the spatial distribution of reward values for a tethered drone approaching a waypoint near the perching target. The perching target is represented by a red point. Dark blue areas indicate restricted zones, where penalties are applied according to \(p_{\text{zone}}\). Individual reward illustrations can be accessed through the GitHub repository.}
    \label{fig:reward-function-heatmap}
\end{figure}

We structure the reward function to provide feedback across three main phases of the perching task: Approaching (I), Wrapping (II) and Hanging (IV), with a focus on I and II. The Descending phase (III) and Hanging phase (IV), though part of the overall perching process, are managed by the attitude controller in real-world experiments and are therefore not the main focus of the reward function. The Hanging phase (IV) serves in reward to inform task completion and reinforce stability in the final position. The entire reward system \( r(\mathbf{o}_t) \) is constructed as follows:
\begin{align}
\small
r(\mathbf{o}_t) =
\begin{cases}
2\mathbf{R}_{\text{approach}} + \mathbf{R}_{\text{wrap}}, & \text{[I] }w(\mathbf{o_t}) \leq 0.5\label{eq:total_reward}\\
 2 + 2\mathbf{R}_{\text{wrap}} + \mathbf{P}_{\text{collision}}, &\text{[II] } w(\mathbf{o}_t) \in \left(\frac{1}{2}, 1\right)\\
2 + 2\mathbf{R}_{\text{wrap}} + \mathbf{P}_{\text{collision}} + \mathbf{R}_{\text{hang}} , &\text{[IV] } w(\mathbf{o}_t) \geq 1.0
\end{cases}
\end{align} where \( w(\mathbf{o}_t) \), the number of wraps, is calculated as:
\begin{equation} \label{eq:wraps}
w(\mathbf{o}_t) = \frac{1}{2\pi} \left| \sum_{t=1}^{T} \Delta \theta^t_{pw} \right|
\end{equation}
with \( \Delta \theta^t_{pw} \) representing the angular change of the perching weight, normalized within \([- \pi, \pi]\). As it is shown, the reward design is not hard-coded for a specific value, such as \( l_{\text{tether}} \) or starting point. 

The \(r(\mathbf{o}_t)\) is consisted of four components: \(\mathbf{R}_{\text{approach}}(\mathbf{o}_t) \), \(\mathbf{R}_{\text{wrap}}(\mathbf{o}_t) \), \(\mathbf{R}_{\text{hang}}(\mathbf{o}_t) \), and \(\mathbf{P}_{\text{collision}}(\mathbf{o}_t) \). The Approaching Reward \( \mathbf{R}_{\text{approach}}(\mathbf{o}_t) \) is illustrated in Figure \ref{fig:reward-function-heatmap}, which encourages effective movement toward the perching target by summing four factors: \(
\mathbf{R}_{\text{approach}}(\mathbf{o}_t)=\tanh(r_{\text{proximity}}+ r_{\text{endwaypoint}}+ r_{\text{tether}}+ p_{\text{zone}})
\). The Wrapping Reward \( \mathbf{R}_{\text{wrap}}(\mathbf{o}_t) \) promotes effective tether wrapping and is defined as:
\begin{equation} \label{eq:wrapping_reward}
\mathbf{R}_{\text{wrap}}(\mathbf{o}_t) = \frac{1}{2}  \left(1 + \tanh\left(2  (w(\mathbf{o}_t) - 1)\right)\right).
\end{equation} The Hanging Reward \( \mathbf{R}_{\text{hang}}(\mathbf{o}_t) \) designed an ideal hanging zone to complete perching task, while the Collision Penalty \( \mathbf{P}_{\text{collision}} \) further reduces the reward for close contact with the perching target during stage II and IV, thereby reducing the collision. The implementation and visualization of \( \mathbf{P}_{\text{collision}} \), and \( \mathbf{R}_{\text{hang}}(\mathbf{o}_t) \) can be found on our open repository. 

As part of the Approaching Reward \( \mathbf{R}_{\text{approach}}(\mathbf{o}_t) \), the Proximity Reward \( r_{\text{proximity}}(\mathbf{o}_t) \) aims to minimize the distance to the perching target and is calculated as:
\begin{equation}\label{eq:proximity_reward}
\small
\begin{aligned}
r_{\text{proximity}}(\mathbf{o}_t)=\;
&\max\!\left(
      -1,\,
      \min\!\left(
          0,\,
          \frac{d_b-C_{\text{threshold}}}{C_{\text{scale}}}
      \right)
\right)\\
&\quad
+\;\tanh\!\left(1-\frac{d_t}{2}\right),
\end{aligned}
\end{equation} where \( d_b \) is the distance from the quadrotor $\mathbf{x}_{\text{q}}$ to the perching target centre $\mathbf{x}_{\text{target}}$, \( d_t \) is the distance to the desired end waypoint, and constants \( C_{\text{threshold}} \) and \( C_{\text{scale}} \) define a threshold for safe distance (1 meter), and penalty constant as the quadrotor deviates from this threshold, respectively. The Desired End Waypoint Proximity Reward \( r_{\text{endwaypoint}}(\mathbf{o}_t) \) further promotes an approach based on \( d_t \), defined as:
\begin{equation} \label{eq:target_reward}
\small
r_{\text{endwaypoint}}(\mathbf{o}_t) = 
\begin{cases}
1.0 & \text{if } d_t < 0.05 \\
0.75 & \text{if } 0.05 \leq d_t < 0.10 \\
0.5 & \text{if } 0.10 \leq d_t < 0.25 \\
0.25 & \text{if } 0.25 \leq d_t < 0.50 \\
0.1 & \text{if } 0.50 \leq d_t < 1.00 \\
0.0 & \text{if } d_t \geq 1.00
\end{cases}
\end{equation}

Moreover, the Tether Contact Reward \( r_{\text{tether}}(\mathbf{o}_t) \) promotes continuous contact with the perching target at phase I, particularly for the final segment, and is defined as:
\begin{equation}
\small
\begin{split}
    r_{\text{tether}}(\mathbf{o}_t) = &\ \delta_{\text{contact}}(t) \cdot \min(C_{\text{max}}, t \cdot C_{\text{increment}}) \\
    &\ + \left(1 - \frac{\max(0, d_\text{tb} - C_{\text{offset}})}{C_{\text{scale}}}\right),
\end{split}
\end{equation} where $d_\text{tb}$ defines the distance between the last one-third of the tether (ideal contact point) $\mathbf{x}_{\mathrm{contact-point}}$ , and the center of the perching target $\mathbf{x}_{\mathrm{target}}$. The term \( \delta_{\text{contact}}(t) \cdot \min(C_{\text{max}}, t \cdot C_{\text{increment}}) \) rewards sustained contact between any part of the tether $\mathcal{L}$  and perching target $\mathcal{P}$, and clears to 0 if contact interrupts, with \( C_{\text{max}} \) as a cap for reward accumulation over time. The second term linearly decreases the reward as \( d_\text{tb} \) exceeds \( C_{\text{offset}} \). 

The Zone Penalty \( p_{\text{zone}}(\mathbf{o}_t) \) discourages entry into restricted zones during approaching:
\begin{equation}
\small
p_{\text{zone}}(\mathbf{o}_t)=
\begin{cases}
d_b - C_{\text{safe}}, &
  \text{if } 
  \theta_{dc}\in[170^{\circ},360^{\circ}]\cup[0^{\circ},10^{\circ}]\\
& \quad\land\; d_b\le C_{\text{safe}}, \\

-(C_{\text{safe}} - d'), &
  \text{if } 
  \theta'_{dc} \in [0^{\circ},180^{\circ}]
  \;\land\;
  d' \le C_{\text{safe}},\\

0, & \text{otherwise}
\end{cases}
\label{eq:zone_penalty}
\end{equation} where \( \theta_{dc} \) and \( \theta'_{dc} \) are angular positions relative to the lower and upper arcs, with \( d' \) the distance from the quadrotor to the centre of the upper arc. Figure \ref{fig:reward-function-heatmap} shows the defined area of arcs. The center of the upper arc is defined as 1.8 meters above the center of the perching target, and the center of the perching target is also the center of the lower arc.  \(C_{\text{safe}}\) defines the threshold distance from the quadrotor to the perching target where the penalty applies.

In terms of the observation space limitation, we set the limit as 5 meters away from the perching target. During training, if the drone is located away from this space, then an early stop would be triggered. As indicated in Figure \ref{fig:reward-function-heatmap}, the reward value maintains the same maximum negative when the quadrotor are outside the defined observation space. Finally, we achieved a smooth and continuous reward function via normalization and scaling, but we neglected any associated odometry errors.

\subsection{Simulation Environment Setup}
\label{subsection:simulation-environtment-setup}

\begin{figure}[!t]
    \centering
    \begin{subfigure}{0.32\linewidth}
        \centering
        \includegraphics[width=\linewidth]{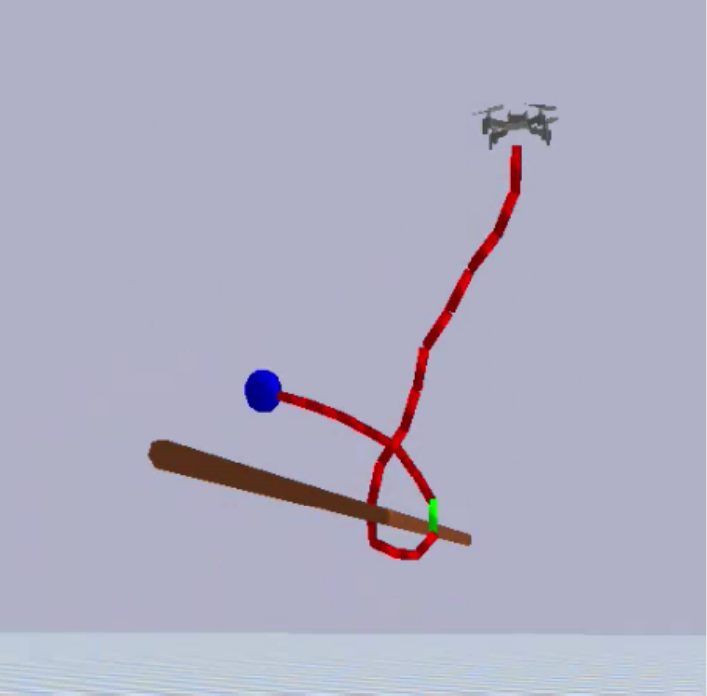}
        \subcaption{}
    \end{subfigure}
    \hfill
    \begin{subfigure}{0.32\linewidth}
        \centering
        \includegraphics[width=\linewidth]{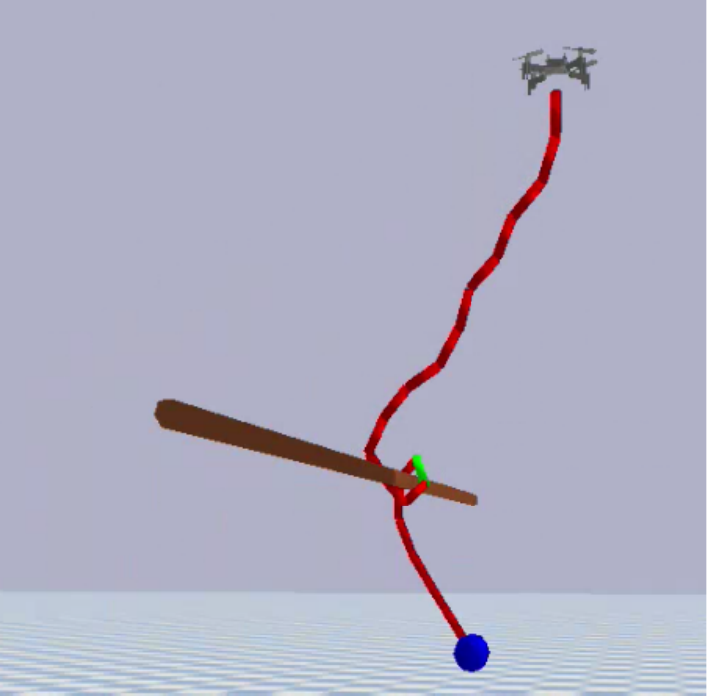}
        \subcaption{}
    \end{subfigure}
    \hfill
    \begin{subfigure}{0.32\linewidth}
        \centering
        \includegraphics[width=\linewidth]{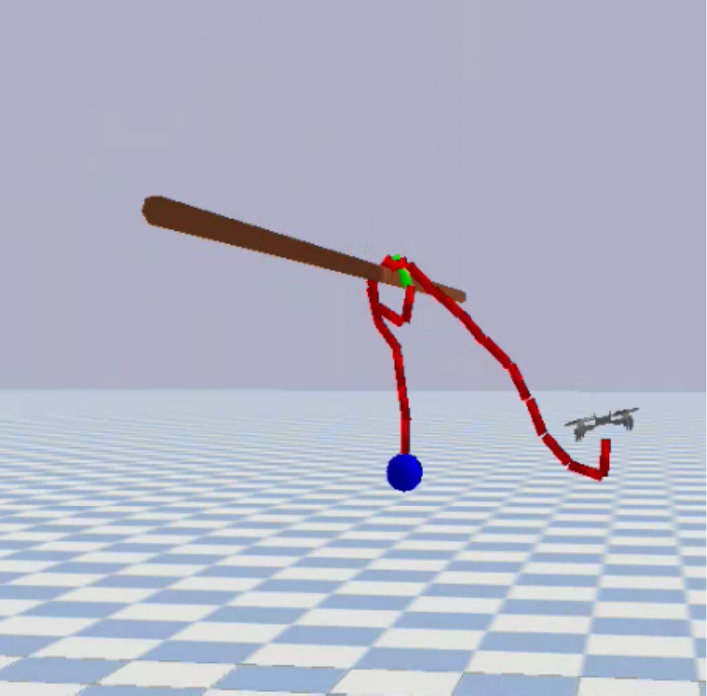}
        \subcaption{}
    \end{subfigure}
    \caption{Simulation phases of aerial robot perching from left to right: (a) initiation of the wrapping, (b) completion of the wrapping, and (c) descending to hanging position.}
    \label{fig:sim-env}
\end{figure}

We set up a simulation using the gym-pybullet-drones framework \cite{panerati_learning_2021}, which provides quadrotor dynamics and a PID controller, and is based on PyBullet, a real-time physics engine that manages collision and constraint enforcement. All the parameters are available through the shared repository. This setup enables realistic interactions between the quadrotor, tensile perching mechanism, and environment, providing effective feedback for training reinforcement learning algorithms. We extended this framework by incorporating a tensile perching mechanism that includes a suspended cable and a perching weight, with a branch-like structure acting as the perching target, as shown in Figure \ref{fig:sim-env}. We performed all simulations and reinforcement learning training on a system equipped with an Intel® Core™ i7-13700H processor, NVIDIA® GeForce RTX™ 4060 GPU (8GB GDDR6), and 32GB DDR5 RAM (2x16GB, 4800MHz).

\subsection{Real-World Experiment Configuration}
\label{section:real-world-experiment-configuration}

We conducted real-world experiments using a 16-camera Vicon T40 motion capture system operating at 100 Hz to provide high-precision positional feedback for the quadrotor platform. This feedback was streamed to the onboard companion computer, which utilized a tracking controller to process the data and ensure accurate trajectory following. The quadrotor was a custom-designed Holybro Kopis Mini 3" racing platform, weighing 230 grams and equipped with a Holybro H7 Mini running PX4 Firmware v1.15.

To enable tensile perching, we integrated a 1-meter-long tether and a 10-gram perching weight attached to the bottom side of the platform. As illustrated in Figure \ref{fig:overview}, the trained RL policy outputs a position path
$\tau_{\mathrm{RL}}=\{\mathbf{x}^{d}_{q,i}\}$ and the off-board computer runs a trajectory optimiser that computed corresponding velocity respecting the hardware limit and kinetic energy required for the tethered perching. This yields the command set
$\tau =\{(\mathbf{x}_{q,i}^{d},\dot{\mathbf{x}}^{d}_{q,i})\}$
(see Figure \ref{fig:whole-sequences}), which is sent over Wi-Fi to the onboard companion computer and relayed via UART to the flight controller for execution.

Upon completion of the wrapping manoeuvre (end of Phase~II in Figure ~\ref{fig:whole-sequences}), in real-world experiments, the control architecture transitions from position control mode to attitude-control mode on PX4, to operate Phase~III and IV. The outer position loop is deliberately disengaged: once the tether is taut, Cartesian error correction would act against the cable constraint and excite oscillations. We retain PX4's cascaded attitude--rate PID but overwrite its set‑points with a constant thrust $T = \lambda T_{\text{hover}}$ (here $\lambda = 0.70$, i.e., \SI{70}{\percent} of hover thrust for the \SI{2.3}{N} weight) and an attitude reference that delivers a desired cable tension $F_{\text{ref}}$. Real‑time tension is estimated as $\hat F = m g - T\cos\phi$ using only commanded thrust and measured tilt; no dedicated force sensor is required. Neglecting aerodynamic drag, vertical force balance gives $F_{\text{ref}} = m g - T\cos\phi_d$, yielding
\begin{equation}\label{eq:attitude-controller}
  \phi_d = \arccos\bigl((m g - F_{\text{ref}})/T\bigr), \qquad |\phi_d| \le 25^{\circ},
\end{equation} which is converted to a quaternion $q_d(\phi_d)$ and sent to PX4 controller.

The tether is held taut for \SI{3}{s}, after which $T$ is ramped linearly from $0.70$ to $0.30\,T_{\text{hover}}$ over the next \SI{4}{s}, giving a total descending phase of \SI{5}{s}, while $\phi_d$ is recomputed from~\eqref{eq:attitude-controller}. Descent terminates when the branch clearance drops below \SI{0.3}{m} or the commanded thrust remains under $0.18\,T_{\text{hover}}$ for \SI{0.5}{s}; then disarms the vehicle and initiates Hanging Phase (IV).

\section{Results and Discussion} \label{chap:expr}

\subsection{Simulation Training and Evaluation}  

\begin{figure}[!t]
    \centering
    \includegraphics[width=0.9\linewidth]{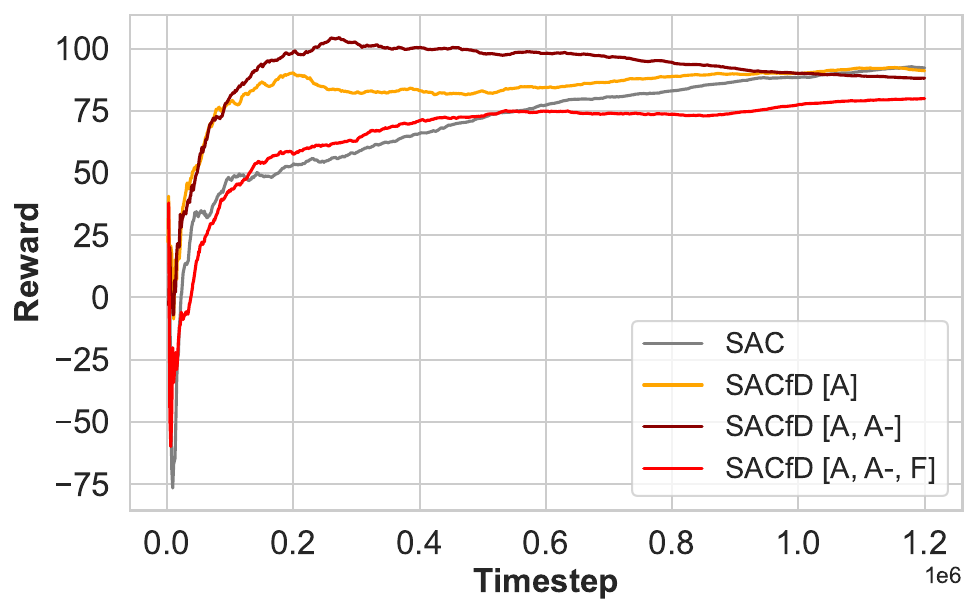}
    \caption{Training performance comparison across various reinforcement learning agents.}
    \label{fig:training-performance}
\end{figure}

\begin{table}[!t]
    \caption{Description of SAC and SACfD Agents.}
    \centering
    \footnotesize
    \begin{tabular}{p{0.12\columnwidth} p{0.4\columnwidth} >{\centering\arraybackslash}p{0.22\columnwidth}}
        \hline
        \textbf{Agent} & \textbf{Description} & \textbf{Total Expert Demonstrations} \\
        \hline
        SAC & Baseline agent trained without demonstrations & None \\
        \hline
        SACfD \(\mathbf{[A]}\)& Agent trained with two optimal demonstrations & 2 \\
        \hline
        SACfD \(\mathbf{[A,A-]}\)& Agent trained with two optimal and three sub-optimal demonstrations & 5 \\
        \hline
        SACfD \(\mathbf{[A,A-,F]}\)& Agent trained with two optimal, three sub-optimal, and one failed demonstration & 6 \\
        \hline
    \end{tabular}
    \label{tab:agent_descriptions}
\end{table}

We evaluated the perching strategies learned by our RL agents, focusing on their performance during the critical phase from hovering to completing the wrapping manoeuvre around a perching target. To investigate the impact of demonstration quality and diversity on performance, We trained four different agents: SAC, SACfD \(\mathbf{[A]}\), SACfD  \(\mathbf{[A,A-]}\), and SACfD \(\mathbf{[A,A-,F]}\) and compared their learning efficiency and control strategies, as summarized in Table \ref{tab:agent_descriptions}.

The SAC agent was trained only through reinforcement learning, with no expert demonstrations, in contrast to the SACfD agents, which were trained using various combinations of expert demonstration sets \(\mathbf{A}\), \(\mathbf{A-}\), and \(\mathbf{F}\). Set \(\mathbf{A}\) consisted of two optimal demonstrations that successfully completed the wrapping manoeuvre. Set \(\mathbf{A-}\) included three sub-optimal demonstrations where the aerial robot made contact with the perching target but failed to complete the wrapping. Set \(\mathbf{F}\) contained one failed demonstration characterized by erratic motion and no successful contact.

We trained all four agents and plotted their learning curves in Figure \ref{fig:training-performance}, which shows the reward progression over 1.2 million timesteps, equivalent to approximately 4000 episodes. The baseline agent, SAC, relied purely on exploration without any demonstrations, which resulted in slower learning and lower initial rewards. In contrast, the SACfD \(\mathbf{[A]}\)  agent, trained using only Set \(\mathbf{A}\), which consisted of two optimal demonstrations, quickly achieved higher rewards.  The SACfD \(\mathbf{[A,A-]}\) agent,  included both optimal (Set \(\mathbf{A}\)) and sub-optimal (Set \(\mathbf{A-}\)) demonstrations. It achieved the highest reward values throughout training, demonstrating a higher quantity of demonstration data and a variety of experiences improved generalization across different scenarios. During observation, this agent provided the most unique trajectory compared to all other agents. Finally, the SACfD \(\mathbf{[A,A-,F]}\) agent incorporated all three sets, adding failed demonstrations from Set \(\mathbf{F}\). This agent has a slower but steady increase during the training. In the final timestep, it converged to the lowest point.

To evaluate the deviation between simulated and real-world trajectories, we first generated a single successful perching trajectory consisted of manually tuned waypoints, then executed it eight times under the same environmental conditions as used in our final experiments.  The average deviation of the same quadrotor path from sim-to-real are summarized in Table \ref{tab:discrepancy-metrics-sim-vs-experiment}, with the Z-axis displaying the highest RMSE of 0.2443 m, suggesting challenges in vertical control. Nevertheless, the results demonstrate that the RL agents effectively generalize from simulation to real-world conditions, while also identifying areas that require further optimization.

\begin{table}[!t]
    \caption{Error Metrics Comparing Simulated and Experimental Quadrotor Trajectories (in meters).}
    \centering
    \footnotesize
    \begin{threeparttable}
        \begin{tabular}{p{0.1\columnwidth} p{0.15\columnwidth} p{0.15\columnwidth} p{0.15\columnwidth} p{0.15\columnwidth}}
        \hline
        \textbf{Metric} & \textbf{Position X (m)} & \textbf{Position Y (m)} & \textbf{Position Z (m)} & \textbf{Total (m)} \\
        \hline
        \textbf{MBE}  & 0.0369  & 0.0142  & -0.0077 & 0.0145 \\
        \textbf{MAE}  & 0.1784  & 0.0356  & 0.2163  & 0.1434 \\
        \textbf{RMSE} & 0.2157  & 0.0401  & 0.2443  & 0.1667 \\
        \textbf{MSE}  & 0.0465  & 0.0016  & 0.0597  & 0.0359 \\
        \hline
        \end{tabular}
        \begin{tablenotes}
            \tiny 
            \item MBE: Mean Bias Error, MAE: Mean Absolute Error, RMSE: Root Mean Square Error, MSE: Mean Square Error.
        \end{tablenotes}
    \end{threeparttable}
    \label{tab:discrepancy-metrics-sim-vs-experiment}
\end{table}

\subsection{Experimental Validation of Learned Policies}  

\begin{figure}[t]
    \centering
    \begin{subfigure}{0.9\linewidth}
        \centering
        \includegraphics[width=\linewidth]{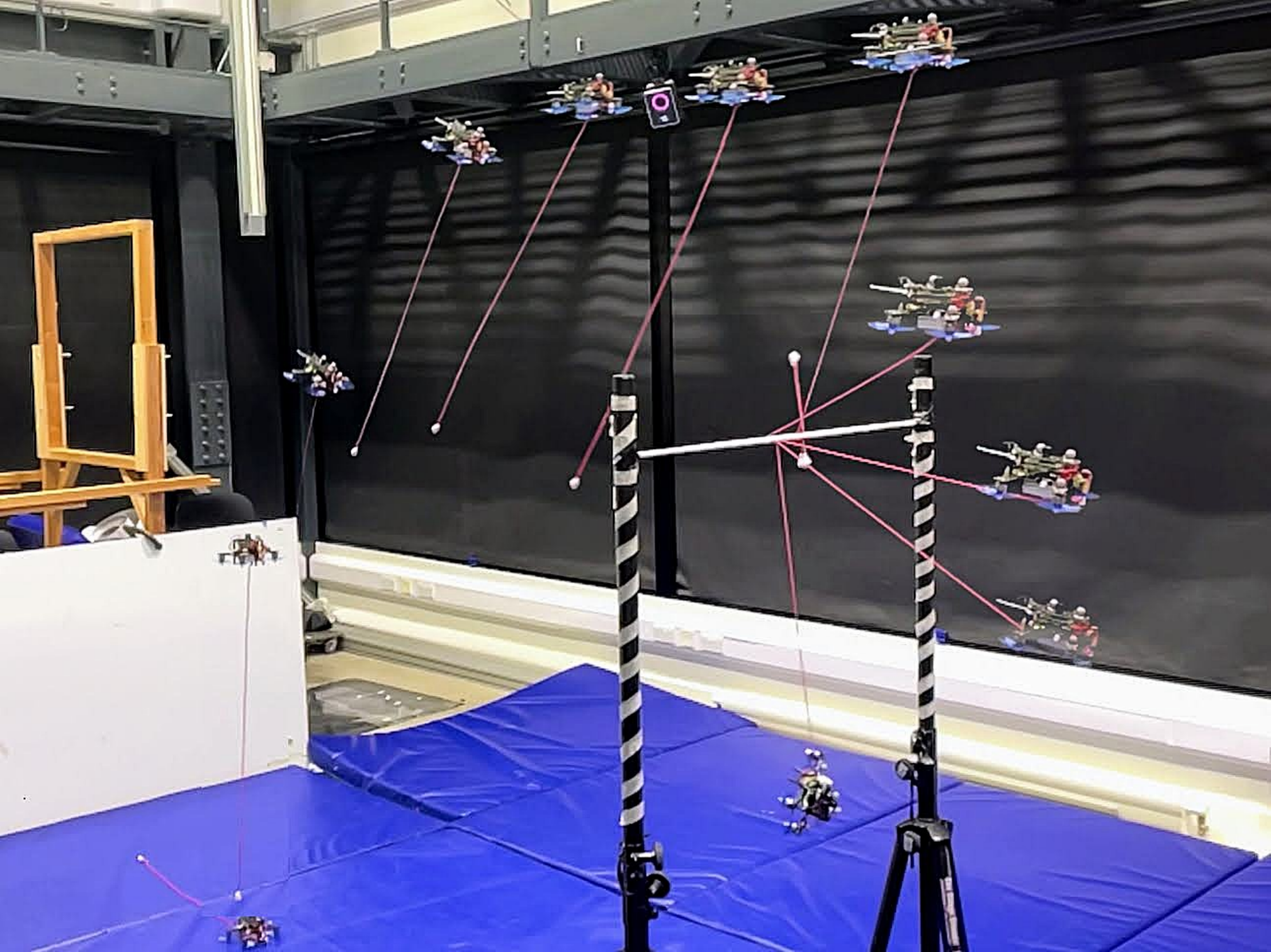}
        \caption{}
        \label{fig:experimental-validation-full-fig}
    \end{subfigure}
    
    \vspace{0.5em} 

    \begin{subfigure}{0.49\linewidth}
        \centering
        \includegraphics[width=\linewidth]{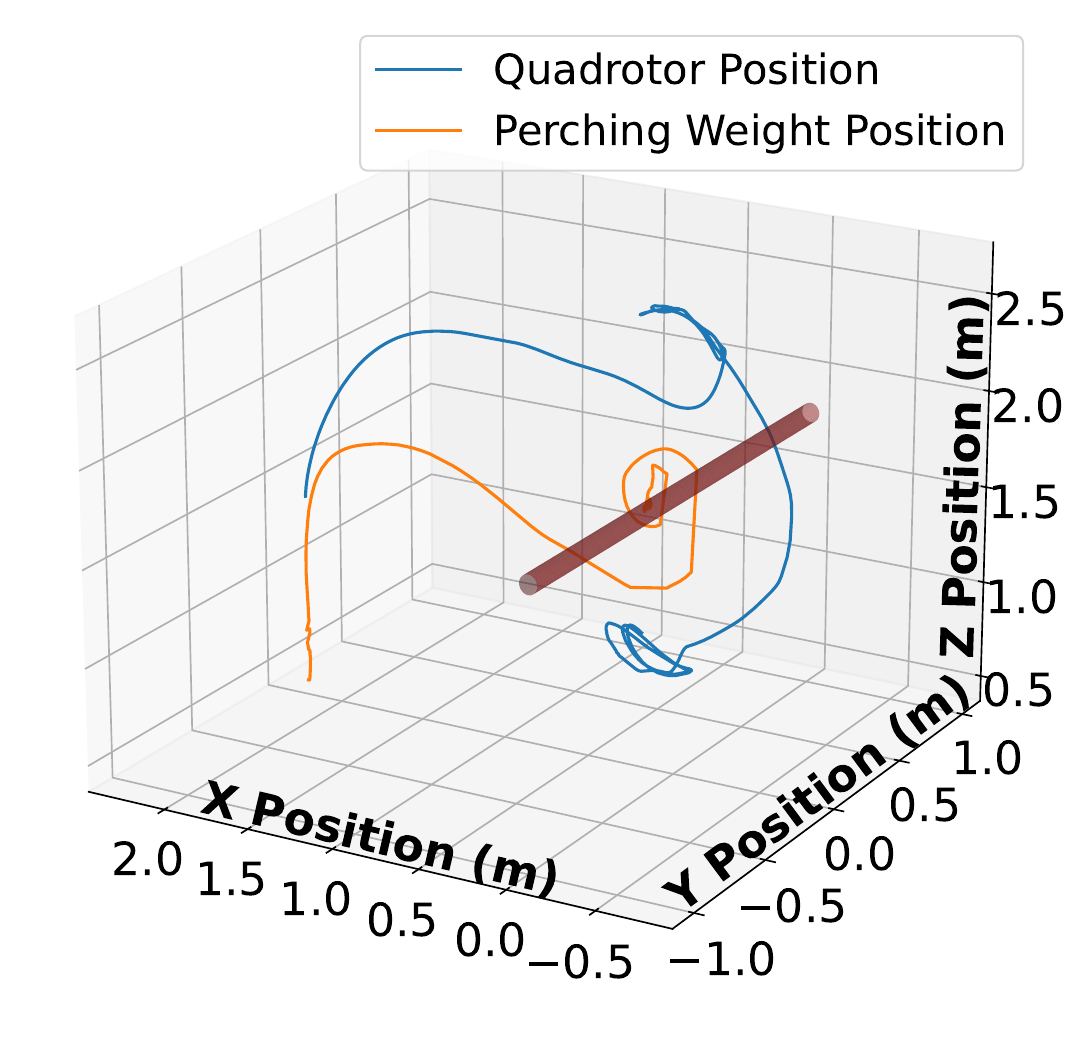}
        
        \caption{}
        \label{fig:example_success_expr}
    \end{subfigure}
    \hfill
    \begin{subfigure}{0.49\linewidth}
        \centering
        \includegraphics[width=\linewidth]{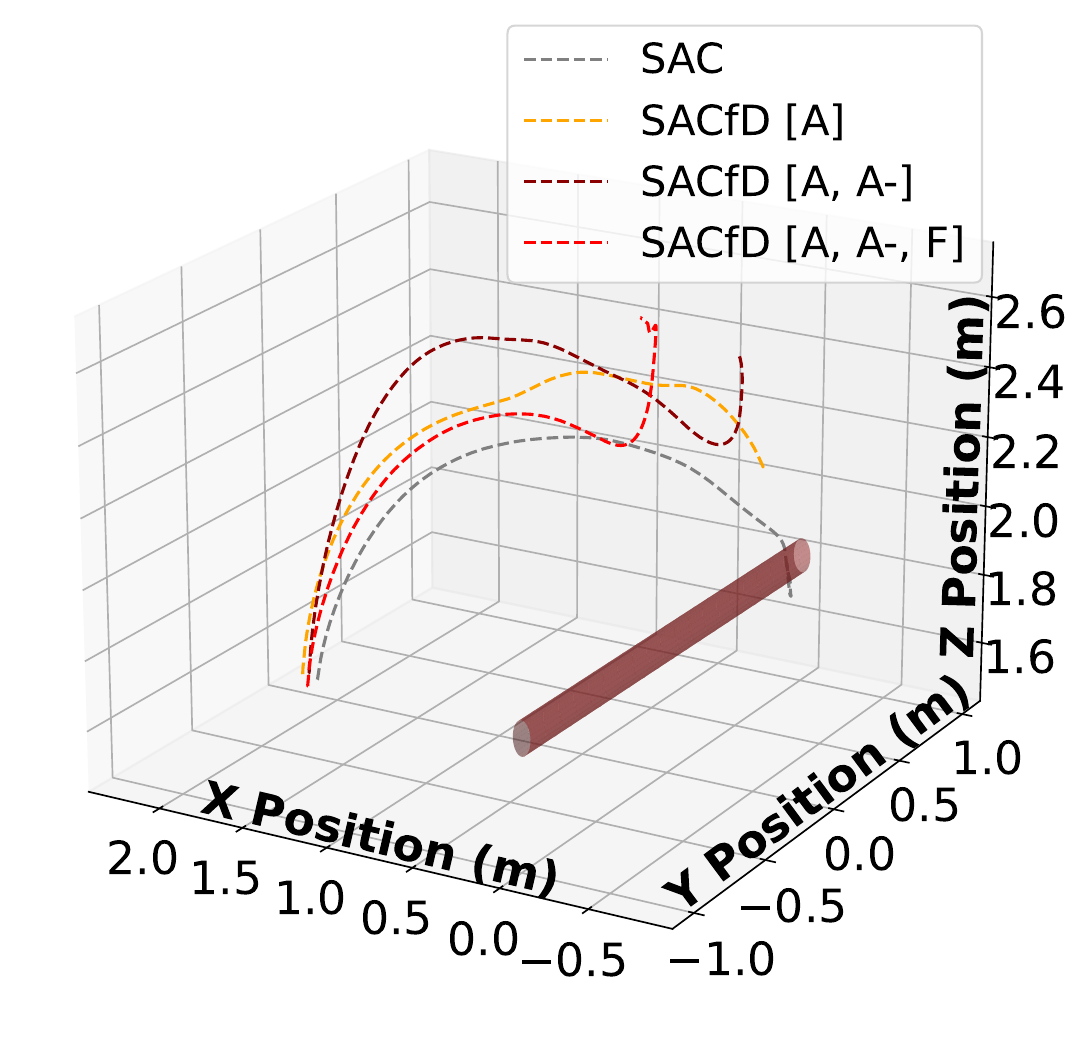}
       
        \caption{}
        \label{fig:overview_4_agents_trajectories}
    \end{subfigure}

    \caption{(a) Aerial robot's full perching sequence, from initial trajectory to stable hanging; (b) Example of a successful perching trajectory from SACfD [\textbf{A, A-}]; (c) Trajectories of four trained agents in real-world experiments.}
    \label{fig:combined_figure}
\end{figure}

We conducted the tethered perching manoeuvre using a quadrotor equipped with a tethered perching system, as depicted in Figure \ref{fig:experimental-validation-full-fig}, and its trajectory of perching weight and quadrotor in \ref{fig:example_success_expr}. In total, 40 flight experiments were performed. These experiments utilized trajectories generated from the best model with the highest reward obtained by four agents during simulation, with each agent contributing 10 experiments. 

To accurately align the target perching location with the simulation (Table \ref{tab:discrepancy-metrics-sim-vs-experiment}), the learned paths were transferred with minimal or no offsets to the onboard computer, where a tracking controller forwarded position and velocity setpoints to the flight controller. Across all 40 trials, conducted with all four agents, every run successfully achieved wrapping and stable hanging, aligning well with the outcomes observed in the simulation. Notably, we tested the perching maneuver with an acceleration peak at approximately 40 \(\text{m/s}^2\), indicating agile and corrective actions to stabilize the quadrotor and ensure successful perching.

\begin{figure}[!t]
    \centering
    \begin{subfigure}{0.49\linewidth}
        \centering
        \includegraphics[width=\linewidth]{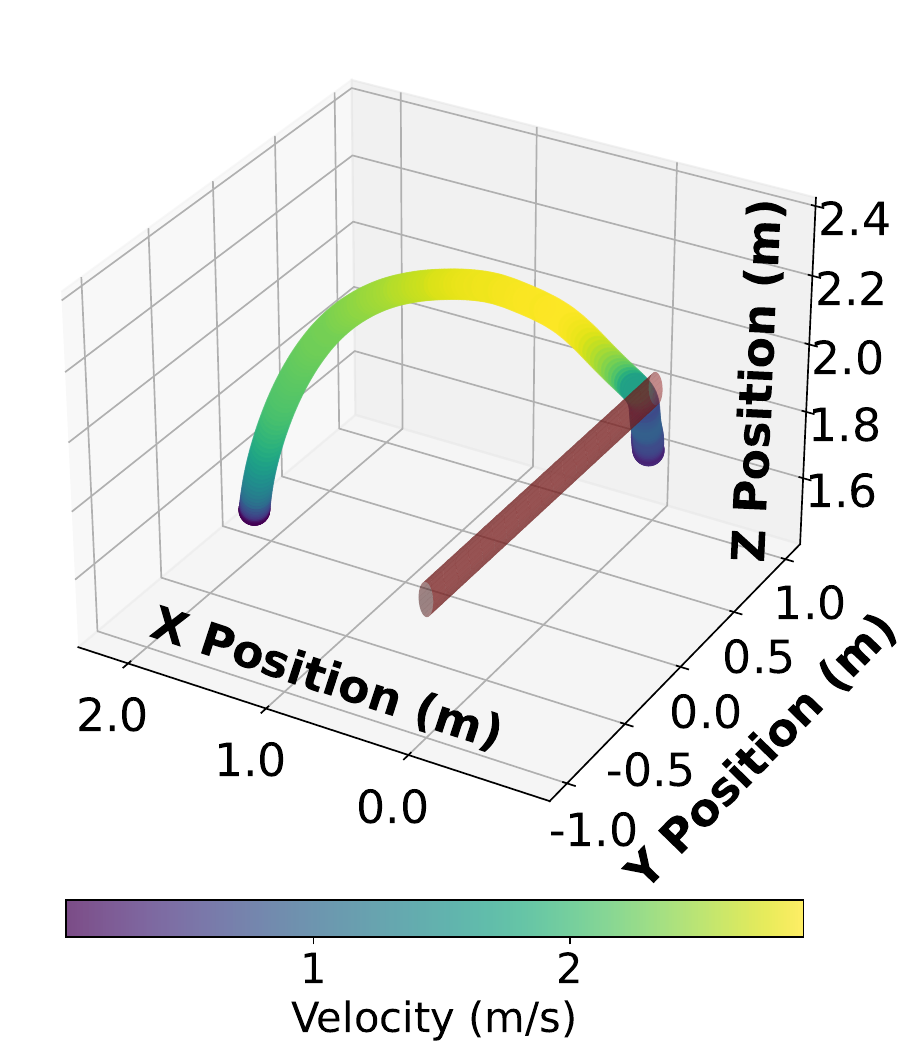}
        \caption{SAC agent}
        \label{fig:sac-agent}
    \end{subfigure}
    \hfill
    \begin{subfigure}{0.49\linewidth}
        \centering
        \includegraphics[width=\linewidth]{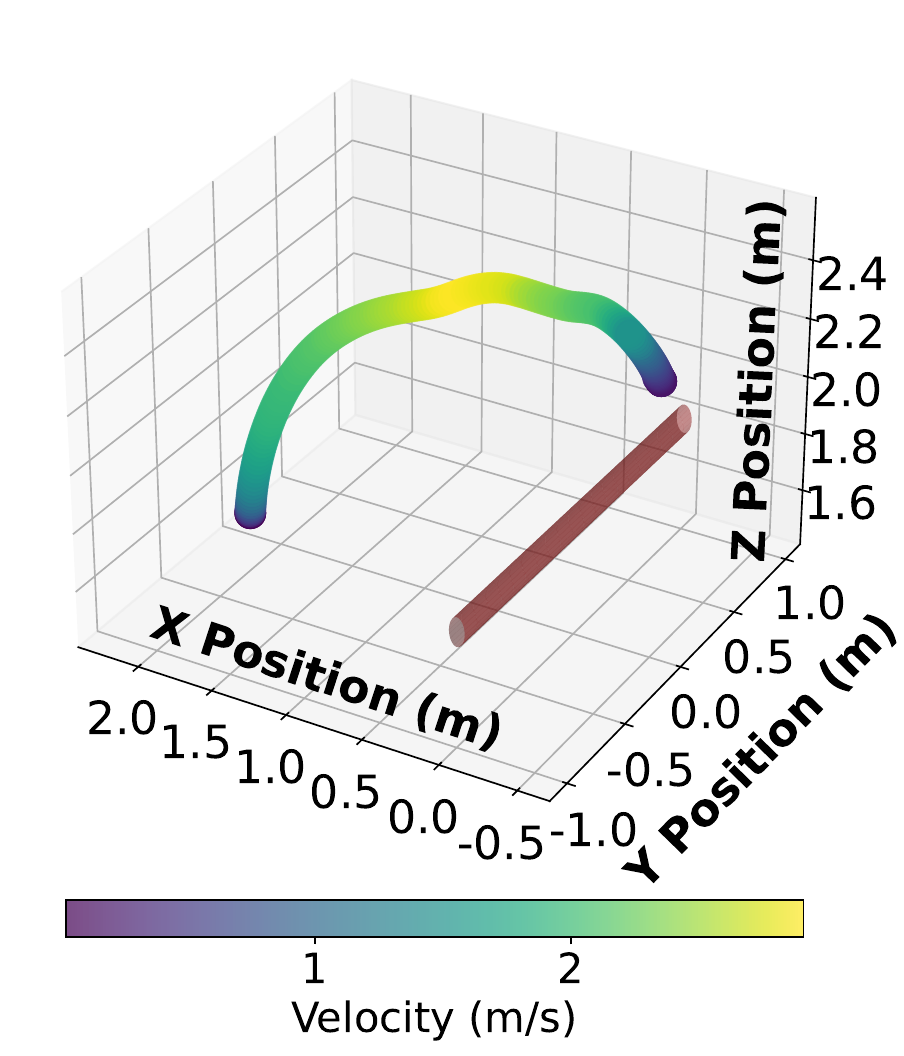}
        \caption{SACfD [\textbf{A}] agent}
        \label{fig:sacfd-agent-a}
    \end{subfigure}
    \hfill
    \begin{subfigure}{0.49\linewidth}
        \centering
        \includegraphics[width=\linewidth]{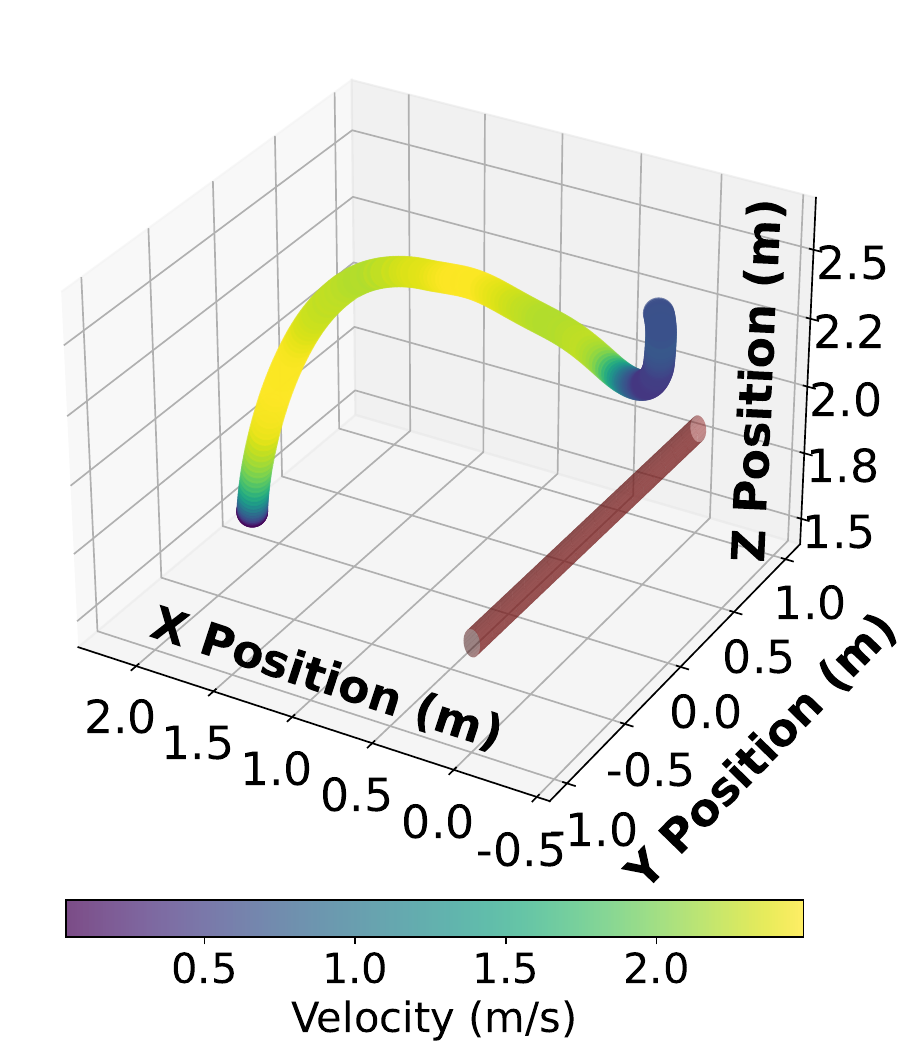}
        \caption{SACfD [\textbf{A,A-}] agent}
        \label{fig:sacfd-agent-ab}
    \end{subfigure}
    \hfill
    \begin{subfigure}{0.49\linewidth}
        \centering
        \includegraphics[width=\linewidth]{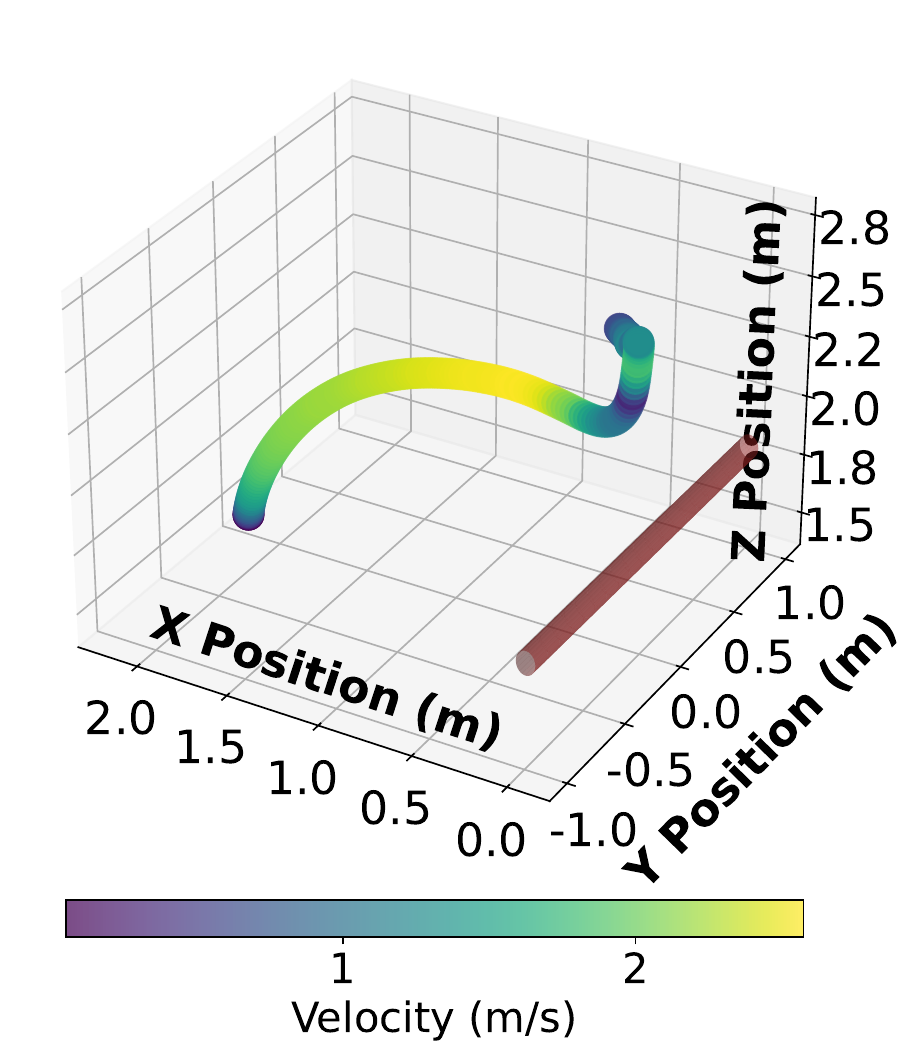}
        \caption{SACfD [\textbf{A,A-,F}] agent}
        \label{fig:sacfd-agent-abc}
    \end{subfigure}

    \caption{Heatmaps of velocity magnitudes during the perching trajectories, showing the velocity distribution of various agents.}
    \label{fig:heatmaps-all-four-trajectories}
\end{figure}

While all agents successfully executed the maneuver, each trajectory exhibited distinct perching strategies. Figure \ref{fig:heatmaps-all-four-trajectories} shows the corresponding trajectories and their velocity distributions for each agent, and Figure \ref{fig:overview_4_agents_trajectories} presents an overview of their learned policy in real-world experiments. Although SACfD \([\mathbf{A}]\) achieved faster convergence in learning from Figure \ref{fig:training-performance}, it produced a trajectory similar to SAC, following a direct and straight perching path. In contrast, SACfD \([\mathbf{A,A-}]\) and SACfD \([\mathbf{A,A-,F}]\) demonstrated different trajectories, featuring a sudden upward movement after reaching the target position. This movement proved to be a beneficial strategy, tightening the cable upon perching impact before descending into a stable hanging position.

Although the current testing shows promising results for the RL controller, perching success is not always guaranteed and can be affected by several factors. During the real-world experiments, we observed an obvious influence of perching velocity, mass of perching weight, and the shape of the perching weight on the final outcome. These results are included in our shared GitHub repository. 




\subsection{Comparison of Learned Policies and Analytical Solution}
\begin{figure}[!t]
    \centering
    \includegraphics[width=1.0\linewidth]{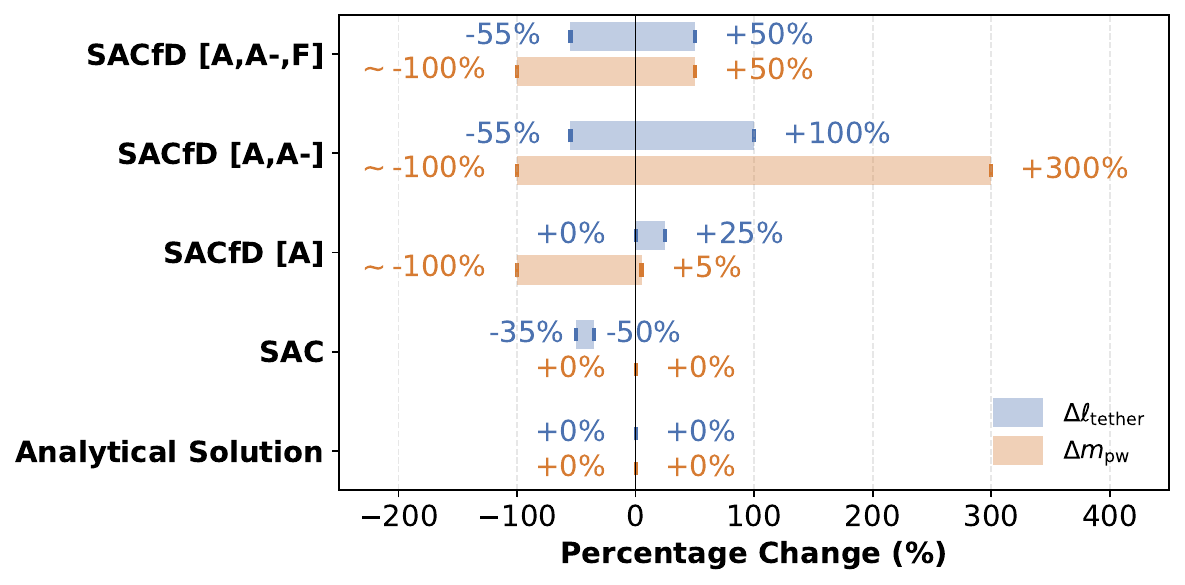}
    \caption{Promising perching range for each agent.  Horizontal bars show the minimum and maximum percentage change relative to nominal value in tether length ($\Delta\ell_{\text{tether}}$) and perching weight mass ($\Delta m_{\text{pw}}$), with promising perching result. The success rate within the displayed range is at least 80\%.}

    \label{fig:safe_range}
\end{figure}
To further validate the RL-Policy training scheme, we conducted comparative experiments on SAC, SACfD \([\mathbf{A}]\),  SACfD \([\mathbf{A, A-}]\), SACfD \([\mathbf{A,A-,F}]\) and the analytical solution from our previous study \cite{hauf_learning_2023}. Each agent runs with an identical start point and perching target position, but varying tether lengths and perching weight masses, so to observe the perching outcomes across different scenarios. All experiments were performed in the same implemented PyBullet simulation. We sample the $m_{\text{pw}}$ and $\ell_{\text{tether}}$ in two sampling tiers. First, we step in \(\pm5\,\%\) increments from \(-100\,\%\) to \(+100\,\%\) relative to the nominal value. Then, we continue upward in \(+20\,\%\) increments to explore the extended range. For each $\ell_{\text{tether}}$ - $m_{\text{pw}}$ combination, we run a batch of five simulation episodes. Success is defined as successful wrapping in the simulation. However, if the failed wrapping is caused by the payload hitting the tether or the quadrotor, the trial is still classified as a success in simulation. The visual appearance of the simulation environment does not always guarantee the perching behavior, we interpret the results according to the real experimental data. The results presented in Figure \ref{fig:safe_range} show the promising perching range of each agent, which was achieved at least 4 successful runs out of 5 episodes per combination. 

Relative to the analytical baseline, whose performance are very unstable across 5 runs with at most 1 trial showing perching, the RL approaches demonstrate greater robustness. The SACfD agents can handle tether length changes of up to -50\% and +100\%, and reach perching weight mass changes up to of -100\% to +300\%. Even though the SAC agent was trained without any demonstration, its performance already surpassed the analytical solution, with tolerance on tether length changes from -35\% to -50\%, and at most a 20\% success rate across different perching weight mass. 

Among the SACfD agents, the SACfD \([\mathbf{A,A-}]\) shows the widest tolerance with 1.5 to 6.2 times more robust than the others in terms of changes in tether length, and 2.7 to 3.8 times wider in perching weight mass changes. This indicates that the involvement of suboptimal demonstrations have advantages over purely optimal demonstrations and failed demonstrations involvement. The 100\% optimal demonstrations in  SACfD \([\mathbf{A}]\) might already perform overfitting, hereby it does not generalize well to wider distribution. The failed demonstrations in the  SACfD \([\mathbf{A,A-,F}]\) might lead the agent to learn the wrong pattern, resulting in downgraded performance. However, compared with the SAC agent without demonstration, the SACfD agents overall improved the tolerance to both the diverse tether length and perching weight mass. Besides, we did not observe conclusive remarks for the energy consumption through accelerations and thrust, since we did not simulate the battery model in our simulation, but we included these results in our open repository. 

\section{Conclusions} \label{chap:conclusion}

In this work, we developed a framework for generating agile trajectories for tensile perching, addressing the unique challenges of tethered aerial perching through reinforcement learning enhanced by demonstration data. Our physics-based simulation environment, combined with dynamic system modelling, served as an effective testbed for training and evaluating our trajectory generation strategies. Using the SACfD algorithm, we derived an optimal trajectory policy that was successfully transferred to a real quadrotor equipped with a tethered perching weight. Experimental results demonstrated that the aerial robot could execute agile and precise tensile perching manoeuvre that closely mirrored the performance achieved in simulation, validating the framework’s efficacy and adaptability in real-world conditions. 

Future work will explore taking off, either by modifying the hardware of the quadrotor, or cutting off the tether after a sensor is placed with the maneuver. This reverse trajectory could be learned or computed.


\section*{Acknowledgment}

The authors thank Tommy Woodley for his initial exploration of the work in an early simulation version. Atar Babgei is supported by Indonesia Endowment Fund for Education (ID 202212220112737).

\flushend

\bibliographystyle{IEEEtran}
\bibliography{new_references}

\begin{thebibliography}{10}
\providecommand{\url}[1]{#1}
\csname url@samestyle\endcsname
\providecommand{\newblock}{\relax}
\providecommand{\bibinfo}[2]{#2}
\providecommand{\BIBentrySTDinterwordspacing}{\spaceskip=0pt\relax}
\providecommand{\BIBentryALTinterwordstretchfactor}{4}
\providecommand{\BIBentryALTinterwordspacing}{\spaceskip=\fontdimen2\font plus
\BIBentryALTinterwordstretchfactor\fontdimen3\font minus \fontdimen4\font\relax}
\providecommand{\BIBforeignlanguage}[2]{{%
\expandafter\ifx\csname l@#1\endcsname\relax
\typeout{** WARNING: IEEEtran.bst: No hyphenation pattern has been}%
\typeout{** loaded for the language `#1'. Using the pattern for}%
\typeout{** the default language instead.}%
\else
\language=\csname l@#1\endcsname
\fi
#2}}
\providecommand{\BIBdecl}{\relax}
\BIBdecl

\bibitem{shakhatreh_unmanned_2019}
H.~Shakhatreh \emph{et~al.}, ``Unmanned {Aerial} {Vehicles} ({UAVs}): {A} {Survey} on {Civil} {Applications} and {Key} {Research} {Challenges},'' \emph{IEEE Access}, vol.~7, pp. 48\,572--48\,634, 2019.

\bibitem{ham_visual_2016}
Y.~Ham \emph{et~al.}, ``Visual monitoring of civil infrastructure systems via camera-equipped unmanned aerial vehicles (uavs): a review of related works,'' \emph{Visualization in Engineering}, vol.~4, pp. 1--8, 2016.

\bibitem{mulero-pazmany_unmanned_2017}
M.~Mulero-Pázmány \emph{et~al.}, ``\BIBforeignlanguage{en}{Unmanned aircraft systems as a new source of disturbance for wildlife: {A} systematic review},'' \emph{\BIBforeignlanguage{en}{PLOS ONE}}, vol.~12, no.~6, p. e0178448, Jun. 2017.

\bibitem{kocer_forest_2021}
B.~B. Kocer \emph{et~al.}, ``Forest drones for environmental sensing and nature conservation,'' in \emph{2021 Aerial Robotic Systems Physically Interacting with the Environment (AIRPHARO)}.\hskip 1em plus 0.5em minus 0.4em\relax IEEE, 2021, pp. 1--8.

\bibitem{nguyen2024crash}
P.~H. Nguyen, ``Crash landing onto" you": Untethered soft aerial robots for safe environmental interaction, sensing, and perching,'' \emph{arXiv preprint arXiv:2405.10043}, 2024.

\bibitem{pritchard2025forestvo}
T.~Pritchard \emph{et~al.}, ``Forestvo: Enhancing visual odometry in forest environments through forestglue,'' \emph{IEEE Robotics and Automation Letters}, vol.~10, no.~6, pp. 5233--5240, 2025.

\bibitem{bates2025leaf}
E.~Bates \emph{et~al.}, ``Leaf level ash dieback disease detection and online severity estimation with uavs,'' \emph{IEEE Access}, vol.~13, pp. 55\,499--55\,511, 2025.

\bibitem{kirchgeorg2024eprobe}
S.~Kirchgeorg \emph{et~al.}, ``eprobe: sampling of environmental dna within tree canopies with drones,'' \emph{Environmental Science \& Technology}, vol.~58, no.~37, pp. 16\,410--16\,420, 2024.

\bibitem{geckeler2025field}
C.~Geckeler \emph{et~al.}, ``Field deployment of biodivx drones in the amazon rainforest for biodiversity monitoring,'' \emph{IEEE Transactions on Field Robotics}, 2025.

\bibitem{zhang_spidermav_2017}
K.~Zhang \emph{et~al.}, ``Spidermav: Perching and stabilizing micro aerial vehicles with bio-inspired tensile anchoring systems,'' in \emph{2017 IEEE/RSJ international conference on intelligent robots and systems (IROS)}.\hskip 1em plus 0.5em minus 0.4em\relax IEEE, 2017, pp. 6849--6854.

\bibitem{braithwaite_tensile_2018}
A.~Braithwaite \emph{et~al.}, ``Tensile web construction and perching with nano aerial vehicles,'' \emph{Robotics Research: Volume 1}, pp. 71--88, 2018.

\bibitem{nguyen_passively_2019}
H.-N. Nguyen \emph{et~al.}, ``A passively adaptive microspine grapple for robust, controllable perching,'' in \emph{2019 2nd IEEE international conference on soft robotics (RoboSoft)}.\hskip 1em plus 0.5em minus 0.4em\relax IEEE, 2019, pp. 80--87.

\bibitem{hauf_learning_2023}
F.~Hauf \emph{et~al.}, ``Learning tethered perching for aerial robots,'' in \emph{2023 IEEE International Conference on Robotics and Automation (ICRA)}.\hskip 1em plus 0.5em minus 0.4em\relax IEEE, 2023, pp. 1298--1304.

\bibitem{lan_aerial_2024}
T.~Lan \emph{et~al.}, ``Aerial tensile perching and disentangling mechanism for long-term environmental monitoring,'' in \emph{2024 IEEE International Conference on Robotics and Automation (ICRA)}.\hskip 1em plus 0.5em minus 0.4em\relax IEEE, 2024, pp. 3827--3833.

\bibitem{zheng_albero_2024}
L.~Zheng \emph{et~al.}, ``Albero: Agile landing on branches for environmental robotics operations,'' \emph{IEEE Robotics and Automation Letters}, vol.~9, no.~3, pp. 2845--2852, 2024.

\bibitem{roderick_bird-inspired_2021}
W.~R. Roderick \emph{et~al.}, ``Bird-inspired dynamic grasping and perching in arboreal environments,'' \emph{Science Robotics}, vol.~6, no.~61, p. eabj7562, 2021.

\bibitem{li2025treecreeper}
H.~Li \emph{et~al.}, ``Treecreeper drone: Adaptive mechanism for passive tree trunk perching,'' \emph{Advanced Intelligent Systems}, p. 2401101.

\bibitem{zou_perch_2023}
Y.~Zou \emph{et~al.}, ``Perch a quadrotor on planes by the ceiling effect,'' in \emph{2023 IEEE 19th International Conference on Automation Science and Engineering (CASE)}.\hskip 1em plus 0.5em minus 0.4em\relax IEEE, 2023, pp. 1--7.

\bibitem{liu_electrically_2023}
H.~Liu \emph{et~al.}, ``Electrically active smart adhesive for a perching-and-takeoff robot,'' \emph{Science Advances}, vol.~9, no.~43, p. eadj3133, 2023.

\bibitem{daler_perching_2013}
L.~Daler \emph{et~al.}, ``A perching mechanism for flying robots using a fibre-based adhesive,'' in \emph{2013 IEEE International Conference on Robotics and Automation}.\hskip 1em plus 0.5em minus 0.4em\relax IEEE, 2013, pp. 4433--4438.

\bibitem{ji_real-time_2022}
J.~Ji \emph{et~al.}, ``Real-time trajectory planning for aerial perching,'' in \emph{2022 IEEE/RSJ International Conference on Intelligent Robots and Systems (IROS)}.\hskip 1em plus 0.5em minus 0.4em\relax IEEE, 2022, pp. 10\,516--10\,522.

\bibitem{garcia-rubiales_magnetic_2019}
F.~Garcia-Rubiales \emph{et~al.}, ``Magnetic detaching system for modular uavs with perching capabilities in industrial environments,'' in \emph{2019 Workshop on Research, Education and Development of Unmanned Aerial Systems (RED UAS)}.\hskip 1em plus 0.5em minus 0.4em\relax IEEE, 2019, pp. 172--176.

\bibitem{palunko_trajectory_2012}
I.~Palunko \emph{et~al.}, ``Trajectory generation for swing-free maneuvers of a quadrotor with suspended payload: A dynamic programming approach,'' in \emph{2012 IEEE international conference on robotics and automation}.\hskip 1em plus 0.5em minus 0.4em\relax IEEE, 2012, pp. 2691--2697.

\bibitem{guerrero_attenuation_2017}
M.~E. Guerrero-S{\'a}nchez \emph{et~al.}, ``Swing-attenuation for a quadrotor transporting a cable-suspended payload,'' \emph{ISA transactions}, vol.~68, pp. 433--449, 2017.

\bibitem{kocer2019aerial}
B.~B. Kocer \emph{et~al.}, ``Aerial robot control in close proximity to ceiling: A force estimation-based nonlinear mpc,'' in \emph{2019 IEEE/RSJ International Conference on Intelligent Robots and Systems (IROS)}.\hskip 1em plus 0.5em minus 0.4em\relax IEEE, 2019, pp. 2813--2819.

\bibitem{lee2021low}
H.~Lee \emph{et~al.}, ``Low-level pose control of tilting multirotor for wall perching tasks using reinforcement learning,'' in \emph{2021 IEEE/RSJ International Conference on Intelligent Robots and Systems (IROS)}.\hskip 1em plus 0.5em minus 0.4em\relax IEEE, 2021, pp. 9669--9676.

\bibitem{ruckin2022adaptive}
J.~R{\"u}ckin \emph{et~al.}, ``Adaptive informative path planning using deep reinforcement learning for uav-based active sensing,'' in \emph{2022 International Conference on Robotics and Automation (ICRA)}.\hskip 1em plus 0.5em minus 0.4em\relax IEEE, 2022, pp. 4473--4479.

\bibitem{rajeswaran2017learning}
A.~Rajeswaran \emph{et~al.}, ``Learning complex dexterous manipulation with deep reinforcement learning and demonstrations,'' \emph{arXiv preprint arXiv:1709.10087}, 2017.

\bibitem{nair2018overcoming}
A.~Nair \emph{et~al.}, ``Overcoming exploration in reinforcement learning with demonstrations,'' in \emph{2018 IEEE international conference on robotics and automation (ICRA)}.\hskip 1em plus 0.5em minus 0.4em\relax IEEE, 2018, pp. 6292--6299.

\bibitem{wang_impact-aware_2024}
H.~Wang \emph{et~al.}, ``Impact-aware planning and control for aerial robots with suspended payloads,'' \emph{IEEE Transactions on Robotics}, vol.~40, pp. 2478--2497, 2024.

\bibitem{nematollahi2022robot}
I.~Nematollahi \emph{et~al.}, ``Robot skill adaptation via soft actor-critic gaussian mixture models,'' in \emph{2022 International Conference on Robotics and Automation (ICRA)}.\hskip 1em plus 0.5em minus 0.4em\relax IEEE, 2022, pp. 8651--8657.

\bibitem{raffin_stable-baselines3_2021}
A.~Raffin \emph{et~al.}, ``Stable-{Baselines3}: {Reliable} {Reinforcement} {Learning} {Implementations},'' \emph{Journal of Machine Learning Research}, vol.~22, no. 268, pp. 1--8, 2021.

\bibitem{panerati_learning_2021}
J.~Panerati \emph{et~al.}, ``Learning to fly—a gym environment with pybullet physics for reinforcement learning of multi-agent quadcopter control,'' in \emph{2021 IEEE/RSJ International Conference on Intelligent Robots and Systems (IROS)}.\hskip 1em plus 0.5em minus 0.4em\relax IEEE, 2021, pp. 7512--7519.

\end{thebibliography}
\end{document}